\documentclass[conference]{templates/ieee/IEEEtran}
\usepackage{mathtools,bm,amsfonts,amssymb,algorithm,algorithmicx,caption,algpseudocode,mathrsfs,multirow,multicol,subcaption,booktabs,threeparttable,balance,tikz,hyperref,breqn,dsfont,upgreek,lineno}

\newcommand{\sysname}{\textsf{OLAS} }
\newcommand{\sysnamens}{\textsf{OLAS}}
\newcommand{\vnet}{\textsf{PredictNet} }
\newcommand{\vnetns}{\textsf{PredictNet}}
\newcommand{\cnet}{\textsf{CoreNet} }
\newcommand{\cnetns}{\textsf{CoreNet}}

\newcommand{\ie}{\emph{i.e.}}
\newcommand{\eg}{\emph{e.g.}}
\newcommand{\etc}{\emph{etc}}

\newtheorem{example}{Example}
\newtheorem{definition}{Definition}
\DeclareMathOperator*{\minimize}{minimize}

\DeclareMathOperator*{\argmin}{argmin}

\makeatletter
\newcommand*\bigcdot{\mathpalette\bigcdot@{.8}}
\newcommand*\bigcdot@[2]{\mathbin{\vcenter{\hbox{\scalebox{#2}{$\m@th#1\bullet$}}}}}
\makeatother

\usepackage{varwidth}
\DeclareCaptionFormat{centercaption}{%
  \begin{varwidth}{\linewidth}%
    \centering
    #1#2#3%
  \end{varwidth}%
}
  
\renewcommand{\vec}[1]{\bm{#1}}
\algnewcommand\algorithmicinput{\textbf{INPUT:}}
\algnewcommand\INPUT{\item[\algorithmicinput]}
\algnewcommand\algorithmicoutput{\textbf{OUTPUT:}}
\algnewcommand\OUTPUT{\item[\algorithmicoutput]}
\algnewcommand\algorithmicforeach{\textbf{for each}}
\algdef{S}[FOR]{ForEach}[1]{\algorithmicforeach\ #1\ \algorithmicdo}

\begin{document}
\bstctlcite{IEEEexample:BSTcontrol}
\selectfont

\title{One-Shot Learning on Attributed Sequences}
\author{
\IEEEauthorblockN{Zhongfang Zhuang, Xiangnan Kong, Elke, Rundensteiner}
\IEEEauthorblockA{
{Worcester Polytechnic Institute} \\
\{zzhuang, xkong, rundenst\}@wpi.edu}
\and
\IEEEauthorblockN{Aditya Arora, Jihane Zouaoui}
\IEEEauthorblockA{{Amadeus IT Group}\\
\{aditya.arora, jihane.zouaoui\}@amadeus.com}
}

\maketitle
\begin{abstract}
    One-shot learning has become an important research topic in the last decade with many real-world applications. 
    The goal of one-shot learning is to classify \textit{unlabeled} instances when there is only one labeled example per class. 
    Conventional problem setting of one-shot learning mainly focuses on the data that is already in a feature space (such as images).
	However, the data instances in real-world applications are often more complex and feature vectors may not be available. 
    In this paper, we study the problem of one-shot learning on attributed sequences, where each instance is composed of a set of attributes (\eg, user profile) and a sequence of categorical items (\eg, clickstream). 
    This problem is important for a variety of real-world applications ranging from fraud prevention to network intrusion detection. 
    This problem is more challenging than the conventional one-shot learning since there are dependencies between attributes and sequences. 
	We design a deep learning framework \sysname to tackle this problem. 
	The proposed \sysname utilizes a twin network to generalize the features 
	from pairwise attributed sequence examples. 
    Empirical results on real-world datasets demonstrate the proposed \sysname can outperform the state-of-the-art methods under a rich variety of parameter settings. 
\end{abstract}

\section{Introduction}
Humans are capable of learning from one, or just a few examples~\cite{lake2011one}, and grasp the patterns. We recognize a person even if we have seen this person's photo only once~\cite{koch2015siamese}. 
Inspired by this capability, one-shot learning, where the goal is to classify previously unseen instances based on \textbf{only one example per class}, has become an important research topic~\cite{fei2003bayesian,fei2006one,miller2000learning}. 

In the literature, conventional approaches to one-shot learning focus on using feature vectors as input in the learning process~\cite{koch2015siamese,wang2012parametric,xing2010brief}, in which each instance is represented as a fixed-size vector (\eg, images). However, data instances in real-world big data applications are often more complex and heterogeneously structured. In this work, we target at one complex data composed of a variable length sequence of categorical items (\eg, a user's clickstream) along with a set of attributes (\eg, a user's profile). We refer to this complex data as \textit{attributed sequences}. Here are two examples of attributed sequences: 
\begin{example}[Network Traffic as Attributed Sequences]
\label{example-traffic}
Network traffic can be modeled as attributed sequences. Namely, it consists of a sequence of packages being sent or received by the routers and a set of attributes indicating the context of the network traffic (\eg, user privileges, security settings, \etc).
\end{example}
\begin{example}[Genes as Attributed Sequences]
    Genes can be represented as attributed sequences, where each gene consists of a DNA \textit{sequence} and a \textit{set} of \textit{attributes} (\eg, PPI, gene ontology, \etc.) indicating the properties of the gene. 
\end{example}
\begin{figure}[t]
    \centering
       \includegraphics[width=\linewidth]{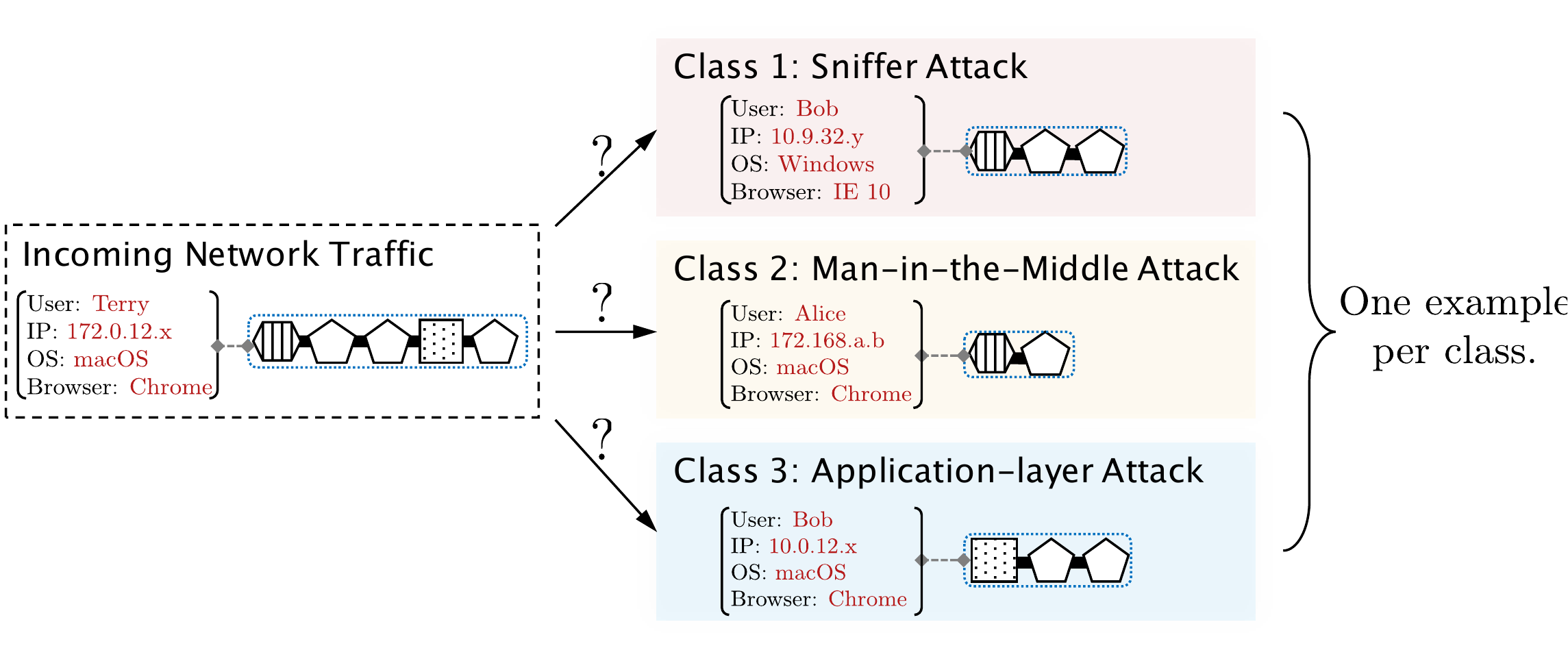}
       \caption{Network attack detection using one-shot learning on attributed sequences. Each instance is composed of a user profile as the \textit{attributes} and a \textit{sequence} of user actions (depicted using different shapes). A system administrator is interested in finding out if the incoming network traffic is malicious with \textbf{only one} sample per class.}
       \vspace{-5mm}
    \label{fig-attributed-sequence}
\end{figure}

Designing one-shot learning to work with attributed sequences promises to be beneficial for a wide range of critical big data applications that require timely responses at scale, such as financial fraud detection and network intrusion detection. In the real-world scenarios, these applications often work on large-scale datasets, yet very few data instances are labeled. 
Continuing with our Example~\ref{example-traffic}, to respond in a timely fashion to potential network intrusion threats, one first has to determine what the intrusion type of incoming potentially malicious traffic even if only one or a few examples per known intrusion type have been seen previously (as depicted in Fig.~\ref{fig-attributed-sequence}). 
Despite its importance in real-world applications, one-shot learning on attributed sequences remains unexplored to date. 

In this paper, we study this new problem of one-shot learning on attributed sequences, with the goal of generating a label for each \textit{unlabeled} attributed sequence with only one training example per known class. This problem is different from previous one-shot learning work, as we now need to extract feature vectors from not only the \textit{attributes} but also the structural information from the \textit{sequences} and the \textit{dependencies} between attributes and sequences. We summarize the specific challenges as follows:
\begin{itemize}
    \item \textbf{Attribute-sequence dependencies.} Fundamental problems arise when learning to classify attributed sequence data. Contrary to the simplifying assumption that the attributes and sequences in these real-world scenarios are independent, various dependencies between them can arise. For example, in network traffic data, one's behavior of \textit{sending/receiving TCP/UDP packets} (\ie, sequences) may depend on the \textit{device type} (\ie, attributes).  Since conventional one-shot learning approaches focus on a single data type, these \textit{dependencies} would thus not be captured. 
    \item \textbf{Generalization in complex data type.} The key difficulty in one-shot learning is to generalize beyond the single training example. It is more difficult to generalize from a more complex data type~\cite{chen2014big}, such as attributed sequence data, than from a simpler data type due to the larger search space and slower convergence. 
\end{itemize}

\textbf{Our Approach.} To address the above challenges, we propose an end-to-end one-shot learning model, called \sysnamens, to accomplish one-shot learning for attributed sequences. The \sysname model includes two main components: a \cnet to encode the information from attributes, sequences and their dependencies and a \vnet to learn the similarities and differences between different attributed sequence classes. The proposed \sysname model is beyond a simple concatenation of \cnet and \vnetns. Instead, they are interconnected within one network architecture and thus can be trained synchronously. Once the \sysname is trained, we can then use it to make predictions for not only the new data but also for entire previously unseen new classes.  
Our paper offers the following core contributions: 
\begin{itemize}
    \item We formulate and analyze the problem of one-shot learning on attributed sequences. 
    \item We develop a deep learning model that is capable of inferring class labels for attributed sequences based on one instance per class. 
    \item We demonstrate that the \sysname network model trained on attributed sequences significantly improves the accuracy of label prediction compared to state-of-the-art methods.
\end{itemize}

We organize the rest as follows. We first define our problem in Section~\ref{section-formulation}. We detail our study of this problem and solve it using a distance metric learning-based solution in Sections~\ref{section-model}. Next, we present the experimental methodology and results in Section~\ref{section-experiments}. We analyze related work in Section~\ref{section-related}. We conclude our findings in Section~\ref{section-conclusion}.

\section{Problem Formulation}
\label{section-formulation}
In this section, we introduce the key definitions and problem formulation of one-shot learning on attributed sequences. The important notations are summarized in Table~\ref{tab-notation}. 
    
\subsection{Preliminaries}
\begin{definition}[Sequence]{\rm
Given a finite set $\mathcal{I}$ composed of $r$ categorical items, a sequence $s_i = \left(x_i^{(1)}, \cdots, x_i^{(t_i)} \right)$ is an ordered list of $t_i$ items, where $\forall x_i^{(t)} \in \mathcal{I}$. }
\end{definition}
The subscript $i$ is used to distinguish different instances. 
One common method for preprocessing variable-length sequences for deep learning is to first \textit{zero-pad} each sequence to the maximum length of the sequences in a dataset, followed by \textit{one-hot encoding} each sequence~\cite{graves2013generating}. We adopt this approach in this work. We denote the maximum length of sequences as $t_{\text{max}}$. Learning models are capable of disregarding the padding so that the padding has no effect in the training of models. 
We denote the one-hot encoded form of sequence $s_i$ as a matrix $\mathbf{s}_i\in \mathbb{R}^{t_{\text{max}} \times r}$. 

\begin{definition}[Attributed Sequence]{\rm
    An attributed sequence $p_i$ is a pair composed of an attribute vector $\mathbf{v}_i$ and a one-hot encoded sequence $\mathbf{s}_i$, denoted as $p_i = (\mathbf{v}_i, \mathbf{s}_i)$. A $u$-dimensional attribute vector $\mathbf{v}_i$ is composed of $u$ attributes in the dataset. }
\end{definition}
\subsection{Problem Definition} 
Inspired by the work in~\cite{bertinetto2016learning}, we formulate our problem as finding the parameters $\theta$ of a predictor $\Theta$ that minimizes the loss $\mathcal{L}_{\text{one-shot}}$. Given a training set of $g$ attributed sequences $\mathcal{G} = \{ (p_1, c_1), \cdots, (p_g, c_g) \}$, where each attributed sequence $p_i$ has a unique class label $c_i$, we formulate the objective for one-shot learning for attributed sequences as:
\begin{equation}
    \label{eq-problem-definition}
    \minimize_\theta \sum_{(p_i, c_i)\in \mathcal{G}} \mathcal{L}_{\text{one-shot}} \left( \Theta\left( p_i  ; \theta  \right), c_i \right)
\end{equation}
That is, we want to minimize the loss calculated using the label predicted using parameter $\theta$ and the true label. 
One-shot learning is known as a hard problem~\cite{koch2015siamese} mainly as a result of unavoidable overfitting caused by insufficient data. With a complex data type, such as attributed sequences, the number of parameters that need to be trained is even larger, which further complicates the problem. 

\begin{table*}[t]
    \centering
    \normalsize
    \caption{Important Mathematical Notations}
    \label{tab-notation}
    \begin{tabular}{cl}
        \hline
        Notation & Description \\ \hline
        $\mathbb{R}$ & The set of real numbers \\
        $r$ & The number of possible items in sequences. \\ 
        $s_i$ & A sequence of categorical items. \\
        $x_i^{(t)}$ & The $t$-th item in sequence $s_i$. \\
        $t_{\text{max}}$ & The maximum length of sequences in a dataset.\\ 
        $\mathbf{s}_i$ & A one-hot encoded sequence in the form of a matrix $\mathbf{s}_i  \in \mathbb{R}^{t_{\text{max}\times r}}$. \\
        $\textbf{x}_i^{(t)}$ & A one-hot encoded item at $t$-th time step in a sequence. \\
        $\mathbf{v}_i$ & An attribute vector. \\
        $p_i$ & An attributed sequence. \ie, $p_i = (\mathbf{v}_i, \mathbf{s}_i)$\\
        $\mathbf{p}_i$ & An $n$-dimensional feature vector of attributed sequence $p_i$. \\
        $\Omega$ & A function transforming each attributed sequence to a feature vector. \\
        $d$ & A distance function. \eg, Mahalanobis distance, Manhattan distance. \\
        $\gamma$ & An activation function within fully connected neural networks. \\ & Possible choices include \texttt{ReLU} and \texttt{tanh}. \\
        $\sigma$ & A logistic activation function within LSTM, \ie, $\sigma(z)=\frac{1}{1+e^{-z}}$ \\
        \hline
    \end{tabular}
\end{table*}

\section{The \textsf{OLAS} Model}
\label{section-model}
\subsection{Approach}
In this work, we adopt an approach from the distance metric learning perspective. Distance metric learning methods are well known for several important applications, such as face recognition, image classification, \etc. Distance metric learning is capable of disseminating data based on their dissimilarities using pairwise training samples. Recent work~\cite{koch2015siamese} has empirically demonstrated the effectiveness of the distance metric learning approach. 
In addition to the pairwise training samples, there are two key components in distance metric learning: a \textit{similarity} label depicting whether the training pair is similar and a distance function $d$. The similar and dissimilar pairs can be randomly generated using the class labels~\cite{koch2015siamese}. We define \textit{attributed sequence triplets} in Definition~\ref{def-triplets}. 
\begin{definition}[Attributed Sequence Triplets]
{\rm \label{def-triplets}
    An attributed sequence triplet $(p_i, p_j, \ell_{ij})$ consists of two attributed sequences $p_i, p_j$, and a \textit{similarity} label $\ell_{ij} \in \{0, 1\}$. The similarity label indicates whether $p_i$ and $p_j$ belong to the same class ($\ell_{ij} = 0$) or different classes ($\ell_{ij} = 1$). We denote $\mathcal{P} = \{(p_i, p_j, \ell_{ij}) | \ell_{ij}=0\}$ as the \textit{positive set} and  $\mathcal{N} = \{(p_i, p_j, \ell_{ij}) | \ell_{ij}=1\}$ as the \textit{negative set}.
    }
\end{definition}
However, attributed sequences are not naturally represented as feature vectors. Therefore, we define a transformation function $\Omega(p_i; \omega)$ parameterized by $\omega$ as a part of the predictor $\Theta$. $\Omega$ uses attributed sequences as the inputs and generates the corresponding feature vectors as the outputs. With two attributed sequences $p_i$ and $p_j$ as inputs, the $n$-dimensional feature vectors of the respective attributed sequences are:
\begin{equation}
\label{eq-two-attseq-vectors}
    \begin{split}        
        \mathbf{p}_i = \Omega(p_i ; \omega)\\
        \mathbf{p}_j = \Omega(p_j ; \omega)\\ 
        \mathbf{p}_i, \mathbf{p}_j \in \mathbb{R}^n
    \end{split}
\end{equation}


The other key component in distance metric learning approaches is a distance function (\eg, Mahalanobis distance ~\cite{cvpr-face-verify}, Manhattan distance~\cite{bertinetto2016learning}). A distance function is applied to the feature vectors in distance metric learning. 

Distance metric learning-based approaches often use the Mahalanobis distance~\cite{cvpr-face-verify,karpathy2015deep}, which can be equivalent to the Euclidean distance~\cite{cvpr-face-verify}. Using the two feature vectors of attributed sequences in Equation~\ref{eq-two-attseq-vectors}, the Mahalanobis distance can be written as: 
\begin{equation}
    \label{eq-mahalanobis-distance}
    d_\omega(\mathbf{p}_i, \mathbf{p}_j) = \sqrt{(\mathbf{p}_i - \mathbf{p}_j)^\top \mathbf\Lambda (\mathbf{p}_i - \mathbf{p}_j)}
\end{equation}
where $d_\omega$ is a specific form of distance function $d$ denoting the inputs (\ie, $\mathbf{p}_i, \mathbf{p}_j$) are the results of transformations using parameter $\omega$. $\mathbf\Lambda \in \mathbb{R}^{n \times n}$ is a symmetric, semi-definite, and positive matrix, and $\mathbf\Lambda$ can be decomposed as:
\begin{equation}
    \mathbf\Lambda = \mathbf\Gamma^\top\mathbf\Gamma,
\end{equation}
where $\mathbf\Gamma \in \mathbb{R}^{f \times n}, f\leq n$. By~\cite{xing2003distance}, Equation~\ref{eq-mahalanobis-distance} is equivalent to:
\begin{equation}
\label{eq-euclidean}
 \begin{split}
    d_\omega(\mathbf{p}_i, \mathbf{p}_j) & = \sqrt{(\mathbf{p}_i - \mathbf{p}_j)^\top \mathbf\Gamma^\top\mathbf\Gamma (\mathbf{p}_i - \mathbf{p}_j)} \\
        &= \| \mathbf\Gamma\mathbf{p}_i- \mathbf\Gamma\mathbf{p}_j\|_2.
 \end{split}
\end{equation}

Instead of directly minimizing the loss of the predictor function $\Theta$ predicting a label of each attributed sequence as in Equation~\ref{eq-problem-definition}, we can now achieve the same training goal by minimizing the loss of predicting whether a pair of attributed sequences belongs to the same class using distance metric learning-based methods. The overall objective can be written as:
\begin{equation}
    \label{eq-problem-def}
    \minimize_{\omega} \sum_{(p_i, p_j, \ell_{ij})\in \mathcal{P}\cup\mathcal{N}} 
    \mathcal{L} \left(
    d_\omega\left(\mathbf{p}_i, \mathbf{p}_j\right), \ell_{ij}
    \right)
\end{equation}
In recent work on distance metric learning applications~\cite{bertinetto2016learning,cvpr-face-verify}, deep neural networks are serve as the nonlinear transformation function $\Omega$. Deep neural networks can effectively learn the features from input data without requiring domain-specific knowledge~\cite{koch2015siamese}, and also generalize the knowledge for future predictions and inferences. These advantages make neural networks become an ideal solution for one-shot learning. 

\subsection{\textsc{OLAS} Model Design}
We next describe the design of the two key components of the \sysname model. First, we design a \cnet for the nonlinear transformation of attributed sequences. Then, a \vnet is designed to learn from the contrast of attributed sequences with different class labels. The specific parameters of the \sysname used in our experiments are detailed in Section~\ref{section-experiments}. 

\begin{figure*}[t]
    \centering
        \includegraphics[width=0.82\linewidth]{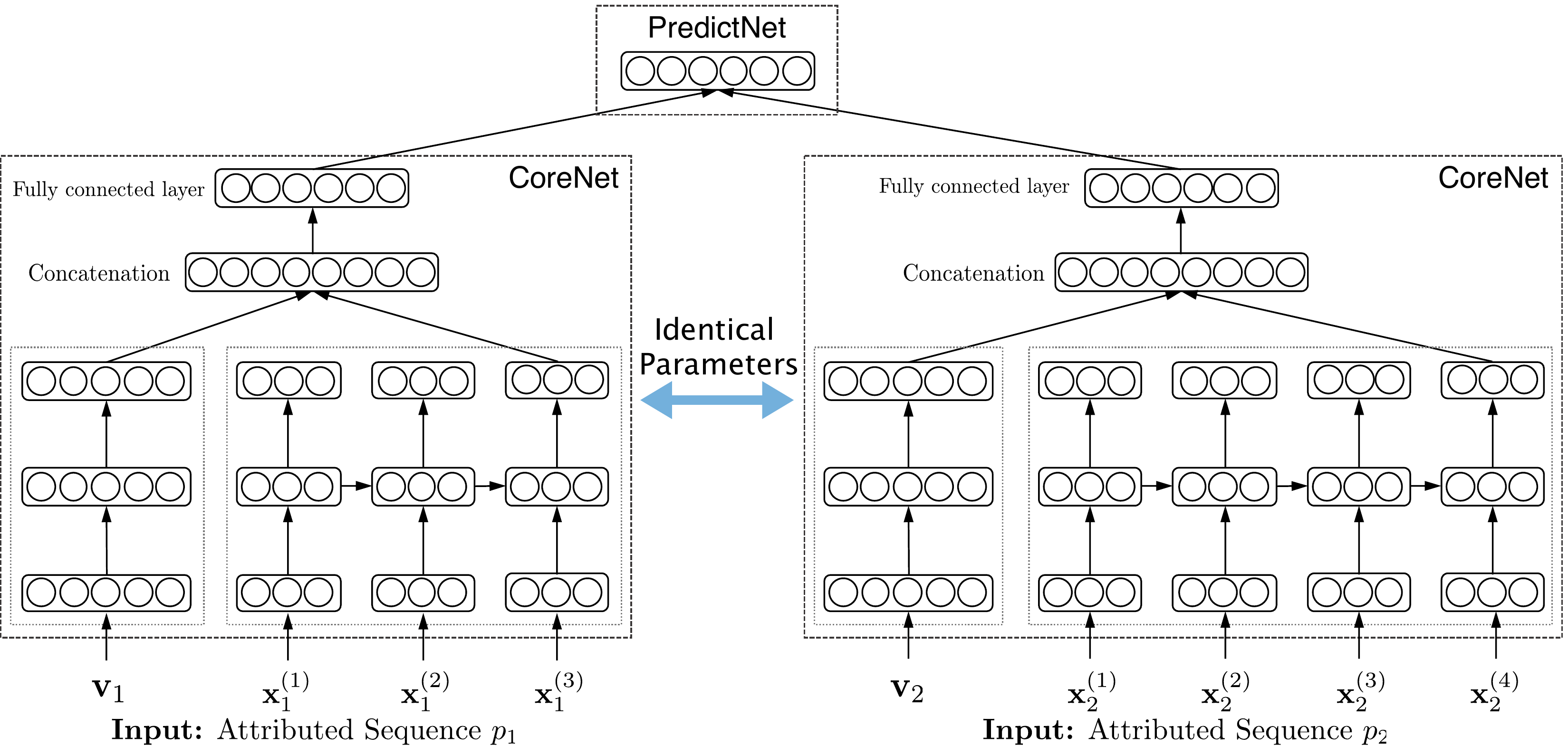}   
        \caption{The network architecture of \sysnamens. The concatenation only happens after the \textit{last time step} of the sequence so the information of the complete sequence is used. }
        \label{fig-fdml}
        \vspace{-5mm}
    \end{figure*}
The two main networks in \cnetns, a fully connected neural network with $m$ layers and a long short-term memory (LSTM) network~\cite{hochreiter1997long}, correspond to the tasks of encoding the information from attributes and sequences in attributed sequences, respectively. By augmenting with another layer of fully connected neural network on top of the concatenation of the above networks, \cnet is also capable of learning the attribute-sequence dependencies.

Given the input of an attribute vector $\mathbf{v}_k\in\mathbb{R}^u$, we define a fully connected neural network with $m$ layers as: 
\begin{equation}
\begin{split}
\label{eq-att-network}
\pmb\upalpha_1 &= \gamma\left(\mathbf{W}_1\mathbf{v}_i + \mathbf{b}_1\right) \\
\pmb\upalpha_2 &= \gamma\left(\mathbf{W}_2\pmb\upalpha_1 + \mathbf{b}_2\right) \\
\vdots \\
\pmb\upalpha_m &= \gamma\left(\mathbf{W}_m\pmb\upalpha_{m - 1} + \mathbf{b}_m\right)
\end{split}
\end{equation}
where $\gamma$ is a nonlinear transformation function. Although we use hyperbolic tangent $\tanh$ in our model, other nonlinear functions such as rectified linear unit (ReLu)~\cite{nair2010rectified} can also be used depending on the empirical results. We denote the weights and bias parameters as: 
\begin{equation}
    \mathbf{W}_{\text{F}} = [\mathbf{W}_{1}, \cdots, \mathbf{W}_{m}]^\top, \mathbf{b}_{\text{F}} = [\mathbf{b}_{1}, \cdots, \mathbf{b}_{m}]^\top%
\end{equation}
Note that the choice of $m$ is task-specific. Although neural networks with more layers are better at learning hierarchical structure in the data, it is also observed that such networks are challenging to train due to the multiple nonlinear mappings that prevent the information and gradient passing along the computation graph~\cite{pham2017column}.

$\mathbf{W}_{\text{F}}$ and $\mathbf{b}_{\text{F}}$ are used to transform the input of each layer to a lower dimension. This transformation is imperative given the often large number of dimensions of attribute vectors in real-world applications. 
Different from attribute vectors, the categorical items in the sequences in attributed sequences obey temporal ordering. The information of sequences is not only in the item values, but more importantly, in the temporal ordering of these items. In this vein, the \cnet utilizes an LSTM network. LSTM is capable of handling not only the ordering of items, but also the dependencies between different items in the sequences. Given a sequence $\mathbf{s}_i$ as the input, we use an LSTM ~\cite{hochreiter1997long} to process each item $\mathbf{x}_k^{(t)}$ in this sequence as: 
\begin{equation}
  \begin{split}
  \label{eq-lstm}
  \mathbf{i}^{(t)} &= \sigma\left(\mathbf{W}_{\text{i}}\mathbf{x}_k^{(t)} + \mathbf{U}_{\text{i}}\mathbf{h}^{(t-1)} + \mathbf{b}_{\text{i}}\right) \\[2pt]
  \mathbf{f}^{(t)} &= \sigma\left(\mathbf{W}_{\text{f}}\mathbf{x}_k^{(t)} + \mathbf{U}_{\text{f}}\mathbf{h}^{(t-1)} + \mathbf{b}_{\text{f}}\right) \\[2pt]
  \mathbf{o}^{(t)} &= \sigma\left(\mathbf{W}_{\text{o}}\mathbf{x}_k^{(t)} + \mathbf{U}_{\text{o}}\mathbf{h}^{(t-1)} + \mathbf{b}_{\text{o}}\right) \\[2pt]
  \mathbf{g}^{(t)} &= \tanh\left(\mathbf{W}_{\text{c}}\mathbf{x}_k^{(t)} + \mathbf{U}_{\text{c}}\mathbf{h}^{(t-1)} + \mathbf{b}_{\text{c}}\right) \\[2pt]
  \mathbf{c}^{(t)} &= \mathbf{f}^{(t)}\odot\mathbf{c}^{(t-1)} + \mathbf{i}^{(t)} \odot \mathbf{g}^{(t)} \\[2pt]
  \mathbf{h}^{(t)} &= \mathbf{o}^{(t)} \odot \tanh\left(\mathbf{c}^{(t)}\right)
  \end{split}
\end{equation}
where $\sigma$ is a sigmoid activation function, $\odot$ denotes the bitwise multiplication, $\mathbf{i}^{(t)}$, $\mathbf{f}^{(t)}$ and $\mathbf{o}^{(t)}$ are the internal gates of the LSTM, $\mathbf{c}^{(t)}$ and $\mathbf{h}^{(t)}$ are the cell and hidden states of the LSTM. Without loss of generality, we denote LSTM kernel parameters $\mathbf{W}_{\text{L}}$, recurrent parameters $\mathbf{U}_{\text{L}}$ and bias parameters $\mathbf{b}_{\text{L}}$ as:
\begin{equation}
    \begin{split}
        \mathbf{W}_{\text{L}} &= [\mathbf{W}_{\text{i}}, \mathbf{W}_{\text{f}}, \mathbf{W}_{\text{o}}, \mathbf{W}_{\text{c}}]^\top \\[-1pt]
        \mathbf{U}_{\text{L}} &= [\mathbf{U}_{\text{i}}, \mathbf{U}_{\text{f}}, \mathbf{U}_{\text{o}}, \mathbf{U}_{\text{c}}]^\top \\[-1pt]
        \mathbf{b}_{\text{b}} &= [\mathbf{b}_{\text{i}}, \mathbf{b}_{\text{f}}, \mathbf{b}_{\text{o}}, \mathbf{b}_{\text{c}}]^\top
    \end{split}
\end{equation}
The attribute vectors and sequences are processed simultaneously and the outputs of both networks are concatenated together. Instead of using the outputs of the LSTM at every time step, we only concatenate the last output from the LSTM to the output of the fully connected neural network so that the complete sequence information is used. 
After that, another layer of fully connected neural network is used to capture the dependencies between attributes and sequences. 
Given the output dimensions of $\pmb\upalpha_m$ and $\mathbf{h}^{(t)}$ as $n_m$ and $n_l$, respectively, the concatenation and the last fully connected layer of \cnet can be written as: 
\begin{equation}
    \normalsize
    \mathbf{p}_i = \gamma\left(\mathbf{W}_{\text{p}}\left( \pmb\upalpha_m \oplus \mathbf{h}^{(t_i)} \right) + \mathbf{b}_{\text{p}}\right)
\end{equation}
where $\oplus$ represents the concatenation of two vectors, $\mathbf{W}_{\text{p}}\in\mathbb{R}^{n\times(n_m+n_{{l}})}$ and $\mathbf{b}_{\text{p}}\in\mathbb{R}^{n}$ denote the weight matrix and bias vector in this fully connected layer for an $n$-dimensional output. In summary, the \cnet can be written as: 
\begin{equation}
\label{eq-func-omega}
\Omega: \left(\mathbb{R}^{u}, \mathbb{R}^{t_{\text{max}} \times r}\right) \mapsto \mathbb{R}^{n}
\end{equation}

\begin{algorithm}[t]
    \normalsize
    \begin{algorithmic}[1]
    \caption{Training using attributed sequence triplets}
    \label{alg-fase}
    \INPUT A positive set $\mathcal{P}$ and a negative set $\mathcal{N}$ of attributed sequence triplets, the number of layers in fully connected neural networks $m$, learning rate $\lambda$, number of iterations $\phi$ and convergence error $\epsilon$. 
    \OUTPUT Parameters of \sysname ($\{\mathbf{W}_{\text{F}}, \mathbf{b}_{\text{F}}, \mathbf{W}_{\text{L}}, \mathbf{U}_{\text{L}}, \mathbf{b}_{\text{L}}\}$).
    \State{Initialize \sysname network.}
    \ForEach{$\phi^{\prime} = 1, \cdots, \phi$} {\Comment{\small $\phi$ is the maximum number of training epochs.}}
        \ForEach{$(p_i, p_j, \ell_{ij}) \in \mathcal{P}\cup\mathcal{N}$}
            \State{$\mathbf{p}_i \leftarrow \Omega(p_i; \omega)$.}
            \State{$\mathbf{p}_j \leftarrow \Omega(p_j; \omega)$.}
            \State{Compute $d_\omega$.} \Comment{Equation~\ref{eq-euclidean}.}
            \State{Compute the loss $\mathcal{L}_{\phi^{\prime}}(\mathbf{p}_i, \mathbf{p}_j, \ell_{ij})$.} \Comment{Equation~\ref{eq-contrastive-loss}.}
            \If{$|\mathcal{L}_{\phi^{\prime}}(\mathbf{p}_i, \mathbf{p}_j, \ell_{ij}) - \mathcal{L}_{\phi^{\prime}-1}(\mathbf{p}_i, \mathbf{p}_j, \ell_{ij})| < \epsilon$} 
                \State{\textbf{break}} \Comment{Early stopping to avoid overfitting. }
            \Else
                \State{Compute $\frac{\partial \mathcal{L}}{\partial d_\omega}, \frac{\partial d_\omega}{\partial \Omega}$. } \Comment{Equation~\ref{eq-l-omega},~\ref{eq-d-omega}.}
                \State{Compute $\nabla \mathcal{L}$.} \Comment{Equation~\ref{eq-nabla-l}.}
                \State{Update network parameters.} \Comment{Equation~\ref{eq-update-all}.}
            \EndIf
        \EndFor
    \EndFor
    \end{algorithmic}
\end{algorithm}
The two outputs of \cnet ($\mathbf{p}_i$ and $\mathbf{p}_j$) are first generated. Then, $\mathbf{p}_i$, $\mathbf{p}_j$ and the similarity label $\ell_{ij}$, are used by the \vnet to learn the similarities and differences between them. 
The \vnet is designed to utilize a contrastive loss function~\cite{hadsell2006dimensionality} so that attributed sequences in different categories are disseminated. 
The contrastive loss function is composed of two parts: a partial loss for the dissimilar pairs and a partial loss for similar pairs. 
The specific form of contrastive loss of \vnet can be written as:
\begin{equation}
\label{eq-contrastive-loss} 
\begin{split}
    \mathcal{L}(\mathbf{p}_i, \mathbf{p}_j, \ell_{ij}) = \underbrace{\frac{1}{2} \ell_{ij} \Big[\max \big( 0, \upxi - d_\omega(\mathbf{p}_i, \mathbf{p}_j)\big)\Big]^2}_{\text{\small Partial loss for \textit{dissimilar} pairs. }} \\[6pt]  +\underbrace{\frac{1}{2}(1-\ell_{ij})d_\omega^2(\mathbf{p}_i, \mathbf{p}_j)}_{\text{\small Partial loss for \textit{similar} pairs}}
\end{split}
\end{equation}
where $\upxi$ is a margin parameter used to prevent the dataset being reduced to a single point~\cite{xing2003distance}. That is, the attributed sequences with $\ell_{ij} = 1$ are only used to adjust the parameters in the transformation function $\Omega$ if the distance between them is larger than $\upxi$. The architecture of \sysname is illustrated in Fig.~\ref{fig-fdml}.

\subsection{OLAS Model Training}
With the contrastive loss $\mathcal{L}$ computed using Equation~\ref{eq-contrastive-loss}, we can now calculate the gradient $\nabla \mathcal{L}$, which is used to adjust parameters in the network as:
\begin{equation}
\label{eq-nabla-l}
    \nabla \mathcal{L} \equiv 
    \left[
        \frac{\partial{\mathcal{L}}}{\partial{\mathbf{W}_{\text{F}}}}, 
        \frac{\partial{\mathcal{L}}}{\partial{\mathbf{b}_{\text{F}}}}, 
        \frac{\partial{\mathcal{L}}}{\partial{\mathbf{W}_{\text{L}}}}, 
        \frac{\partial{\mathcal{L}}}{\partial{\mathbf{U}_{\text{L}}}}, 
        \frac{\partial{\mathcal{L}}}{\partial{\mathbf{b}_{\text{L}}}}
    \right]
\end{equation}
With the transformation function $\Omega$ and distance function $d$, the explicit form of $\nabla \mathcal{L}$ can be written as:
\begin{equation}
\nabla \mathcal{L} = 
\frac{\partial{\mathcal{L}}}{\partial{d_\omega}}\frac{\partial d_\omega}{\partial \Omega}
 \left[
     \frac{\partial\pmb{\upalpha}_m}{\partial{\mathbf{W}_{\text{F}}}}, 
     \frac{\partial \pmb{\upalpha}_m}{\partial{\mathbf{b}_{\text{F}}}}, 
     \frac{\partial \mathbf{h}^{(t_i)}}{\partial{\mathbf{W}_{\text{L}}}}, 
     \frac{\partial \mathbf{h}^{(t_i)}}{\partial{\mathbf{U}_{\text{L}}}}, 
     \frac{\partial \mathbf{h}^{(t_i)}}{\partial{\mathbf{b}_{\text{L}}}}
\right]
\end{equation}
where
\begin{equation}
\label{eq-l-omega}
\begin{split}
    \frac{\partial \mathcal{L}}{\partial d_\omega} = - \ell_{ij}\max(0, \upxi-&d_\omega(\mathbf{p}_i, \mathbf{p}_j))\\ &+ (1-\ell_{ij})d_\omega(\mathbf{p}_i, \mathbf{p}_j) 
\end{split}
\end{equation}
\begin{equation}
\label{eq-d-omega}
\frac{\partial d_\omega}{\partial \Omega} = \left(\mathbf{p}_i - \mathbf{p}_j\right) \cdot \left(\mathds{1}-(\mathbf{p}_i - \mathbf{p}_j)\right)
\end{equation}
where $\mathds{1}$ is a vector filled with ones. 

We present the derivation of \sysname update functions in our network in Appendix~\ref{app-a}. 
With the learning rate $\lambda$, the parameters $\mathbf{W}_{\text{F}}, \mathbf{W}_{\text{L}}, \mathbf{U}_{\text{L}}, \mathbf{b}_{\text{F}}$ and $\mathbf{b}_{\text{L}}$ can be updated by the following equation until convergence is achieved: 
\begin{equation}
\label{eq-update-all}
    \begin{split}
        \mathbf{W}_{\text{F}} &= \mathbf{W}_{\text{F}} - \lambda  \frac{\partial \mathcal{L}}{\partial \mathbf{W}_{\text{F}}} \\[3pt]
        \mathbf{b}_{\text{F}} &= \mathbf{b}_{\text{F}} - \lambda  \frac{\partial \mathcal{L}}{\partial \mathbf{b}_{\text{F}}} \\[3pt]
        \mathbf{W}_{\text{L}} &= \mathbf{W}_{\text{L}} - \lambda  \frac{\partial \mathcal{L}}{\partial \mathbf{W}_{\text{L}}} \\[3pt]
        \mathbf{U}_{\text{L}} &= \mathbf{U}_{\text{L}} - \lambda  \frac{\partial \mathcal{L}}{\partial \mathbf{U}_{\text{L}}} \\[3pt]
        \mathbf{b}_{\text{L}} &= \mathbf{b}_{\text{L}} - \lambda  \frac{\partial \mathcal{L}}{\partial \mathbf{b}_{\text{L}}} \\[3pt]
    \end{split}
\end{equation}
We summarize the algorithms for updating the \sysname network in Algorithm~\ref{alg-fase}. 

\subsection{Labeling Attributed Sequences}
Once we have trained the \sysname network to recognize the similarities and dissimilarities between exemplars of attributed sequence pairs. The \sysname is then ready to be used to assign labels to unlabeled attributed sequences in one-shot learning. 
Given a test attributed sequence $p_k$ from a set $\mathcal{K}$ of unlabeled instances, a set $\mathcal{G}=\{p_g\}_{g=1}^{G}$ of attributed sequences with $G$ categories, in which there is only one instance per category, and the goal is to classify $p_k$ into one of $G$ categories. 
We can now use the \sysname network with only one forward pass to calculate the distance between $p_k$ with each of the $G$ attributed sequences and the label of the instance that is closest to $p_k$ is then assigned as the label of $p_k$. This process can be defined using maximum similarity as:
\begin{equation}
    \widehat{c_k} = \argmin_g d_\omega(\mathbf{p}_k, \mathbf{p}_g)
\end{equation}
where $\widehat{c_k}$ is the predicted label of $\mathbf{p}_k$. 
%
%

\section{Experiments}
\label{section-experiments}
\subsection{Datasets}
Our solution has been motivated in part by use case scenarios observed at Amadeus related to attributed sequences. For this reason, we now work with the log files of an Amadeus~\cite{amadeus} internal application. 
Also, we apply our methodology to real-world, public available Wikispeedia data~\cite{west2009wikispeedia}. We summarize the data descriptions as follows:

\begin{itemize}
    \item \textbf{Amadeus data (AMS1$\sim$AMS6)}. We sampled six datasets from the log files of an internal application at Amadeus IT Group. Each attributed sequence is composed of a user profile containing information (\eg, system configuration, office name) and a sequence of function names invoked by web click activities (\eg, login, search) ordered by time. 
    \item \textbf{Wikispeedia data (WS1$\sim$WS6)}. Wikispeedia is an online game requiring participants to click through from a given start page to an end page using fewest clicks~\cite{west2009wikispeedia}. We select the \textit{finished} path and extract several properties of each path (\eg, the category of the start path, time spent per click). We also sample six datasets from Wikispeedia. The Wikispeedia data is available through the Stanford Network Analysis Project\footnote{https://snap.stanford.edu/data/wikispeedia.html}~\cite{SNAPwiki}.  
\end{itemize}

Following the protocols in recent work~\cite{koch2015siamese}, we utilize the attributed sequences associated with 60\% of categories to generate attributed sequence triplets and use them in training.  

The class labels used in training and one-shot learning are disjoint sets. Similar to the strategy in~\cite{lake2013one}, where the authors designed a 20-way classification task that attempts to match an alphabet with one of the twenty possible classes, we randomly select one instance in the one-shot learning set and attempt to give it a correct label. We selected 2000 instances for each set used in one-shot learning and compute the accuracy. 
We summarize the number of classes in Table~\ref{tab-classes}.
\begin{table}[t]
    \centering
    \normalsize
    \caption{Number of Classes in Datasets}
    \label{tab-classes}
    \begin{tabular}{c|c|c}
    \hline
    Dataset & Training & One-shot Learning \\ 
    \hline
    AMS1, WS1 & 6 & 4 \\
    AMS2, WS2 & 12 & 8 \\
    AMS3, WS3 & 18 & 12 \\
    AMS4, WS4 & 24 & 16 \\ 
    AMS5, WS5 & 30 & 20 \\
    AMS6, WS6 & 36 & 24 \\
    \hline 
    \end{tabular}
\end{table}


\subsection{Compared Methods} We focus on one-shot learning methods on different data types. We summarize the compared methods in Table~\ref{tab-methods}. 
\begin{table}[t]
    \centering
    \normalsize
    \caption{Compared Methods}
    \label{tab-methods}
    \begin{tabular}{c|c|c}
        \hline
        Name & Data Used & Note \\ \hline
        \textsf{OLAS} & Attributed Sequences & This Work\\ 
        \textsf{OLASEmb} & {\scriptsize Attributed Sequence Embeddings} & This work + \cite{NAS-submission}\\
        \textsf{ATT} & Attributes Only & \cite{koch2015siamese}\\
        \textsf{SEQ} & Sequence Only & \cite{sutskever2014sequence} + \cite{koch2015siamese}\\
        \hline
    \end{tabular}
    \vspace{-7mm}
\end{table}
Specifically, we compare the performance of the following one-shot learning methods: 
\begin{itemize}
    \item \textsf{OLAS}: We first evaluate our proposed method using attributed sequences data. 
    \item \textsf{OLASEmb}: Instead of using the attributed sequence instances as input, we use the embeddings of attributed sequences as the input. 
    We want to find out whether a simpler heuristic combination of state-of-the-art would achieve better performance.  
    \item \textsf{ATT}: This is the state-of-the-art method~\cite{koch2015siamese} using only attributes of the data. 
    \item \textsf{SEQ}: We combine the state-of-the-art in one-shot learning~\cite{koch2015siamese} with sequence-to-sequence learning~\cite{sutskever2014sequence} to be able to utilize sequences in one-shot learning.  
\end{itemize}

\begin{algorithm}[t]
    \normalsize
    \begin{algorithmic}[1]
    \caption{One-shot learning for attributed sequences. }
    \label{alg-olas}
    \INPUT Trained networks \cnet $\Omega$ and \vnetns, a set of unlabeled attributed sequences $\mathcal{K}$, a set of labeled attributed sequence with one example per class $\mathcal{G}$ and a distance function $d$.
    \OUTPUT A set of labeleled attributed sequences $\mathcal{K}^{\prime}$.
    \State{$\mathcal{K}^{\prime} \leftarrow \emptyset$}
    \ForEach{$p_k \in \mathcal{K}$}
    \State{$\varepsilon \leftarrow +\infty$ \Comment{Set initial minimum distance to $+\infty$}}
        \State{$\mathbf{p}_k \leftarrow \Omega(p_k; \omega)$}
        \ForEach{$(p_g, c_g) \in \mathcal{G}$}
            \State{$\mathbf{p}_g \leftarrow \Omega(p_g; \omega)$}
            \If{$d{(\mathbf{p}_k, \mathbf{p}_g)} \le \varepsilon$ }
                \State{$\varepsilon \leftarrow d_\omega{(\mathbf{p}_k, \mathbf{p}_g)}$\Comment{Using \vnetns.}}
                \State{$\widehat{c_k} \leftarrow c_g$} \Comment{Assign the same label of $p_g$ to $p_k$.}
            \EndIf
        \State{$\mathcal{K}^{\prime} \leftarrow (p_k, \widehat{c_k})$} 
        \EndFor
    \EndFor
    \State{\Return $\mathcal{K}$}
    \end{algorithmic}
\end{algorithm}
\begin{figure*}[t]
    \centering
    \includegraphics[width=0.55\textwidth]{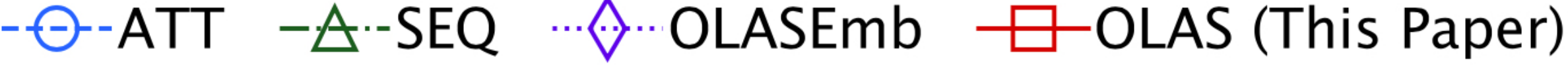}
    
    \begin{subfigure}[t]{0.32\linewidth}
        \centering
        \includegraphics[width=\textwidth, page=1]{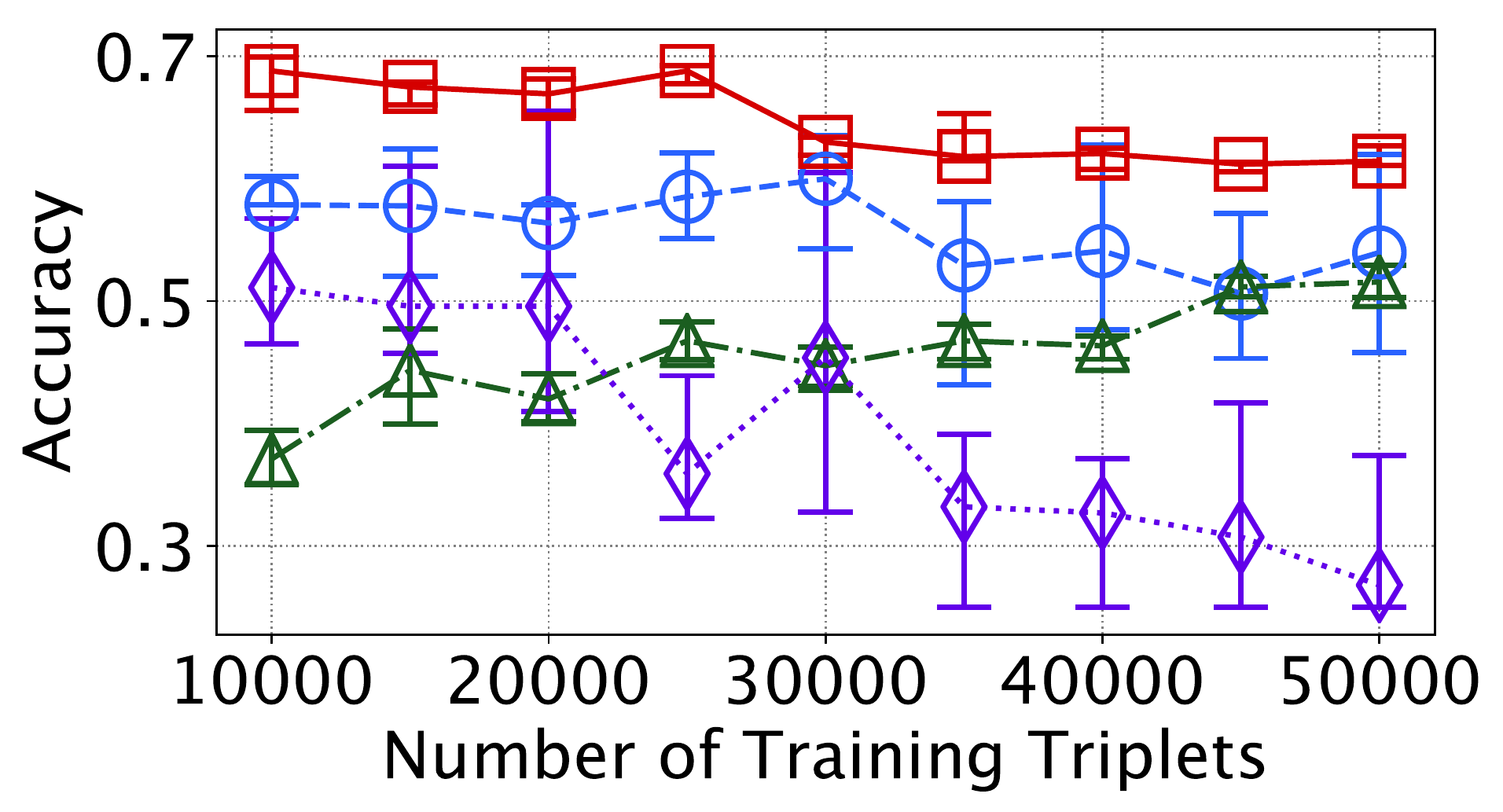}
        \vspace{-6mm}
        \caption{Dataset AMS1}
        \label{fig-exp-euc-ams-1}
    \end{subfigure}
    \begin{subfigure}[t]{0.32\linewidth}
        \centering
        \includegraphics[width=\textwidth, page=2]{figures/experiments/0816/euclidean_deviation}
        \vspace{-6mm}
        \caption{Dataset AMS2}
        \label{fig-exp-euc-ams-2}
    \end{subfigure}
    \begin{subfigure}[t]{0.32\linewidth}
        \centering
        \includegraphics[width=\textwidth, page=3]{figures/experiments/0816/euclidean_deviation}
        \vspace{-6mm}
        \caption{Dataset AMS3}
        \label{fig-exp-euc-ams-3}
    \end{subfigure}
    
    \begin{subfigure}[t]{0.32\linewidth}
        \centering
        \includegraphics[width=\textwidth, page=4]{figures/experiments/0816/euclidean_deviation}
        \vspace{-6mm}
        \caption{Dataset AMS4}
        \label{fig-exp-euc-ams-4}
    \end{subfigure}
    \begin{subfigure}[t]{0.32\linewidth}
        \centering
        \includegraphics[width=\textwidth, page=5]{figures/experiments/0816/euclidean_deviation}
        \vspace{-6mm}
        \caption{Dataset AMS5}
        \label{fig-exp-euc-ams-5}
    \end{subfigure}
    \begin{subfigure}[t]{0.32\linewidth}
        \centering
        \includegraphics[width=\textwidth, page=6]{figures/experiments/0816/euclidean_deviation}
        \vspace{-6mm}
        \caption{Dataset AMS6}
        \label{fig-exp-euc-ams-6}
    \end{subfigure}
    \vspace{-2mm}
    \caption{Accuracy of the label prediction on AMS datasets using \textbf{Euclidean} distance function. }
    \vspace{-2mm}
    \label{fig-exp-euc-ams}
\end{figure*}
\begin{figure*}[t]
    \centering
    \begin{subfigure}[t]{0.32\linewidth}
        \centering
        \includegraphics[width=\textwidth, page=1]{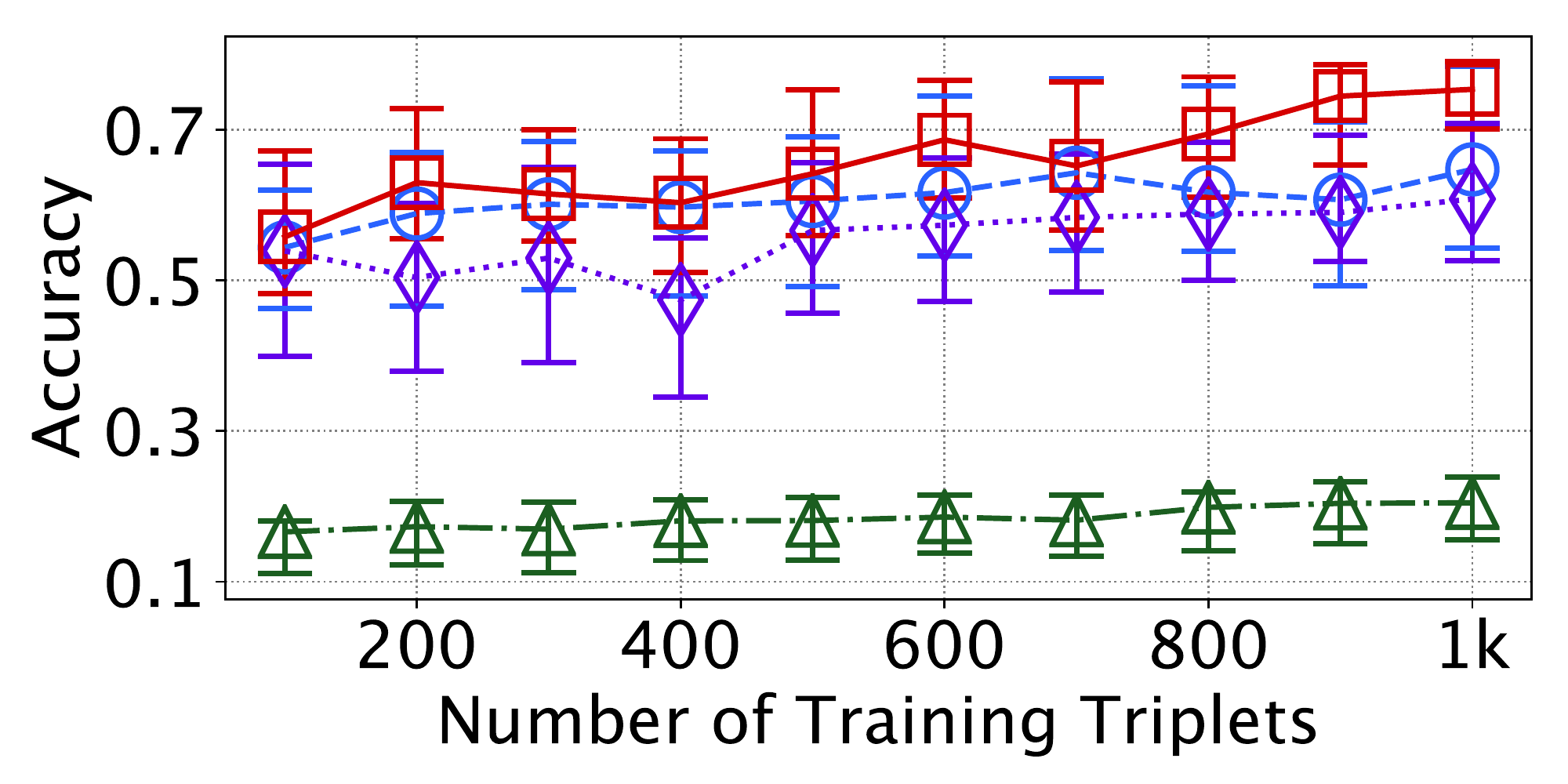}
        \vspace{-6mm}
        \caption{Dataset WS1}
        \label{fig-exp-euc-ws-1}
    \end{subfigure}
    \begin{subfigure}[t]{0.32\linewidth}
        \centering
        \includegraphics[width=\textwidth, page=2]{figures/experiments/0816/ws_euclidean_deviation}
        \vspace{-6mm}
        \caption{Dataset WS2}
        \label{fig-exp-euc-ws-2}
    \end{subfigure}
    \begin{subfigure}[t]{0.32\linewidth}
        \centering
        \includegraphics[width=\textwidth, page=3]{figures/experiments/0816/ws_euclidean_deviation}
        \vspace{-6mm}
        \caption{Dataset WS3}
        \label{fig-exp-euc-ws-3}
    \end{subfigure}
    
    \begin{subfigure}[t]{0.32\linewidth}
        \centering
        \includegraphics[width=\textwidth, page=4]{figures/experiments/0816/ws_euclidean_deviation}
        \vspace{-6mm}
        \caption{Dataset WS4}
        \label{fig-exp-euc-ws-4}
    \end{subfigure}
    \begin{subfigure}[t]{0.32\linewidth}
        \centering
        \includegraphics[width=\textwidth, page=5]{figures/experiments/0816/ws_euclidean_deviation}
        \vspace{-6mm}
        \caption{Dataset WS5}
        \label{fig-exp-euc-ws-5}
    \end{subfigure}
    \begin{subfigure}[t]{0.32\linewidth}
        \centering
        \includegraphics[width=\textwidth, page=6]{figures/experiments/0816/ws_euclidean_deviation}
        \vspace{-6mm}
        \caption{Dataset WS6}
        \label{fig-exp-euc-ws-6}
    \end{subfigure}
    \vspace{-2mm}
    \caption{Accuracy of the label prediction on Wikispeedia datasets using \textbf{Euclidean} distance function. }
    \vspace{-5mm}
    \label{fig-exp-euc-ws}
\end{figure*}
\subsection{Experiment Settings} 
\subsubsection{Protocols} The goal of one-shot learning is to correctly assign class labels to each instance. In order to compare with state-of-the-art work~\cite{koch2015siamese,bertinetto2016learning}, we also use accuracy to evaluate the performance. 
A higher accuracy score means a method could make more correct class label predictions. For each experiment setting, we repeat ten times and report the median, 25 percentile and 75 percentile of the results using error bars. For each training process using attributed sequence triplets, we hold out 20\% of the training data as the validation set. The holdout portion is not limited to the instances with certain labels, but instead, they are randomly chosen from all possible classes. 
\subsubsection{Network Initialization and Settings} 
Gradient-based methods often require a careful initialization of the neural networks. In our experiments, we use normalized random distribution~\cite{glorot2010understanding} to initialize weight matrices $\mathbf{W}_{\text{F}}$ and $\mathbf{W}_{\text{L}}$, orthogonal matrix is used to initialize recurrent matrices $\mathbf{U}_{\text{L}}$ and biases are initialized to zero vector $\pmb 0$. 
 Specifically, the $m$-th layer of the fully connected neural network is initialized as:
 \begin{equation*}
     \mathbf{W}_m \sim \text{Uniform}\left[-\frac{\sqrt{6}}{\sqrt{{n}_{m-1} + {n}_{m+1}}}, \frac{\sqrt{6}}{\sqrt{{n}_{m-1} + {n}_{m+1}}}\right]
 \end{equation*}
 where $n_{m}$ is the output dimension of the $m$-th layer. There are three layers used in our experiments. Meanwhile, the weight matrices $\mathbf{W}_{\text{i}},\mathbf{W}_{\text{f}},\mathbf{W}_{\text{o}},\mathbf{W}_{\text{c}}$ are initialized as:  
 \begin{equation*}
     \mathbf{W}_{\text{i}},\mathbf{W}_{\text{f}},\mathbf{W}_{\text{o}},\mathbf{W}_{\text{c}} \sim \text{Uniform}\left[-\sqrt{\frac{6}{n_l}}, \sqrt{\frac{6}{n_l}}\right]
 \end{equation*}
 where $n_l$ is the dimension of the output. In our experiments, we use 50 dimensions for both $n_l$ and $n_m$.
We utilize $\ell_2$-regularization with early stopping to avoid overfitting. The validation set is composed of 20\% of the total amount of attributed sequence triplets in the training set.  
\subsection{Performance Studies}
In this section, we present the performance studies of the proposed \sysname network and compare it with techniques in the state-of-the-art. 

\begin{figure*}[t]
    \centering
    \includegraphics[width=0.55\textwidth]{figures/experiments/legend}
    
    \begin{subfigure}[t]{0.32\linewidth}
        \centering
        \includegraphics[width=\textwidth, page=1]{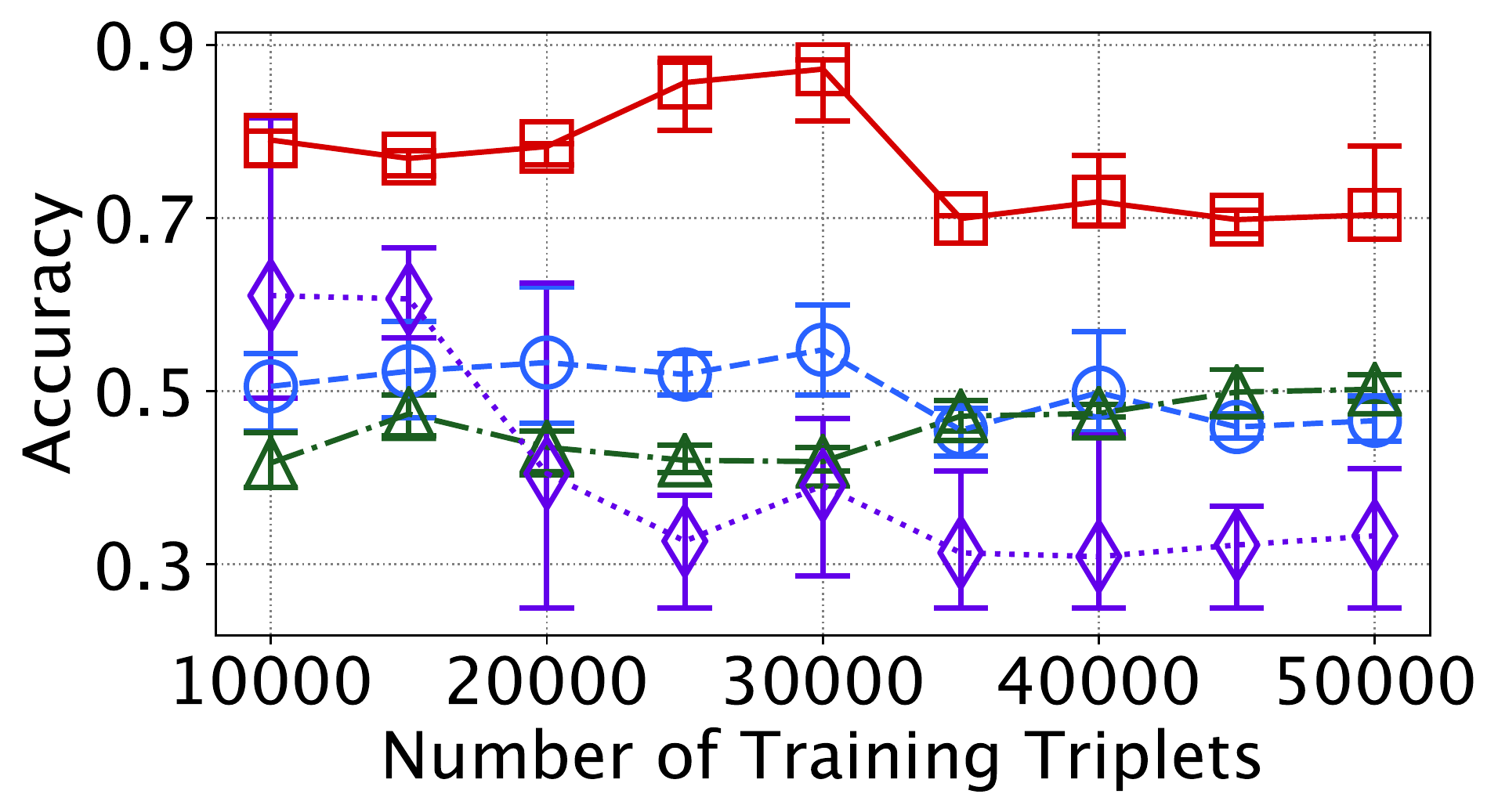}
        \vspace{-6mm}
        \caption{Dataset AMS1}
        \label{fig-exp-mah-ams-1}
    \end{subfigure}
    \begin{subfigure}[t]{0.32\linewidth}
        \centering
        \includegraphics[width=\textwidth, page=2]{figures/experiments/0816/manhattan_deviation}
        \vspace{-6mm}
        \caption{Dataset AMS2}
        \label{fig-exp-mah-ams-2}
    \end{subfigure}
    \begin{subfigure}[t]{0.32\linewidth}
        \centering
        \includegraphics[width=\textwidth, page=3]{figures/experiments/0816/manhattan_deviation}
        \vspace{-6mm}
        \caption{Dataset AMS3}
        \label{fig-exp-mah-ams-3}
    \end{subfigure}
    
    \begin{subfigure}[t]{0.32\linewidth}
        \centering
        \includegraphics[width=\textwidth, page=4]{figures/experiments/0816/manhattan_deviation}
        \vspace{-6mm}
        \caption{Dataset AMS4}
        \label{fig-exp-mah-ams-4}
    \end{subfigure}
    \begin{subfigure}[t]{0.32\linewidth}
        \centering
        \includegraphics[width=\textwidth, page=5]{figures/experiments/0816/manhattan_deviation}
        \vspace{-6mm}
        \caption{Dataset AMS5}
        \label{fig-exp-mah-ams-5}
    \end{subfigure}
    \begin{subfigure}[t]{0.32\linewidth}
        \centering
        \includegraphics[width=\textwidth, page=6]{figures/experiments/0816/manhattan_deviation}
        \vspace{-6mm}
        \caption{Dataset AMS6}
        \label{fig-exp-mah-ams-6}
    \end{subfigure}
    \vspace{-2mm}
    \caption{Accuracy of the label prediction on AMS datasets using \textbf{Manhattan} distance function. }
    \vspace{-3mm}
    \label{fig-exp-mah-ams}
\end{figure*}
\begin{figure*}[t]
    \centering
    \begin{subfigure}[t]{0.32\linewidth}
        \centering
        \includegraphics[width=\textwidth, page=1]{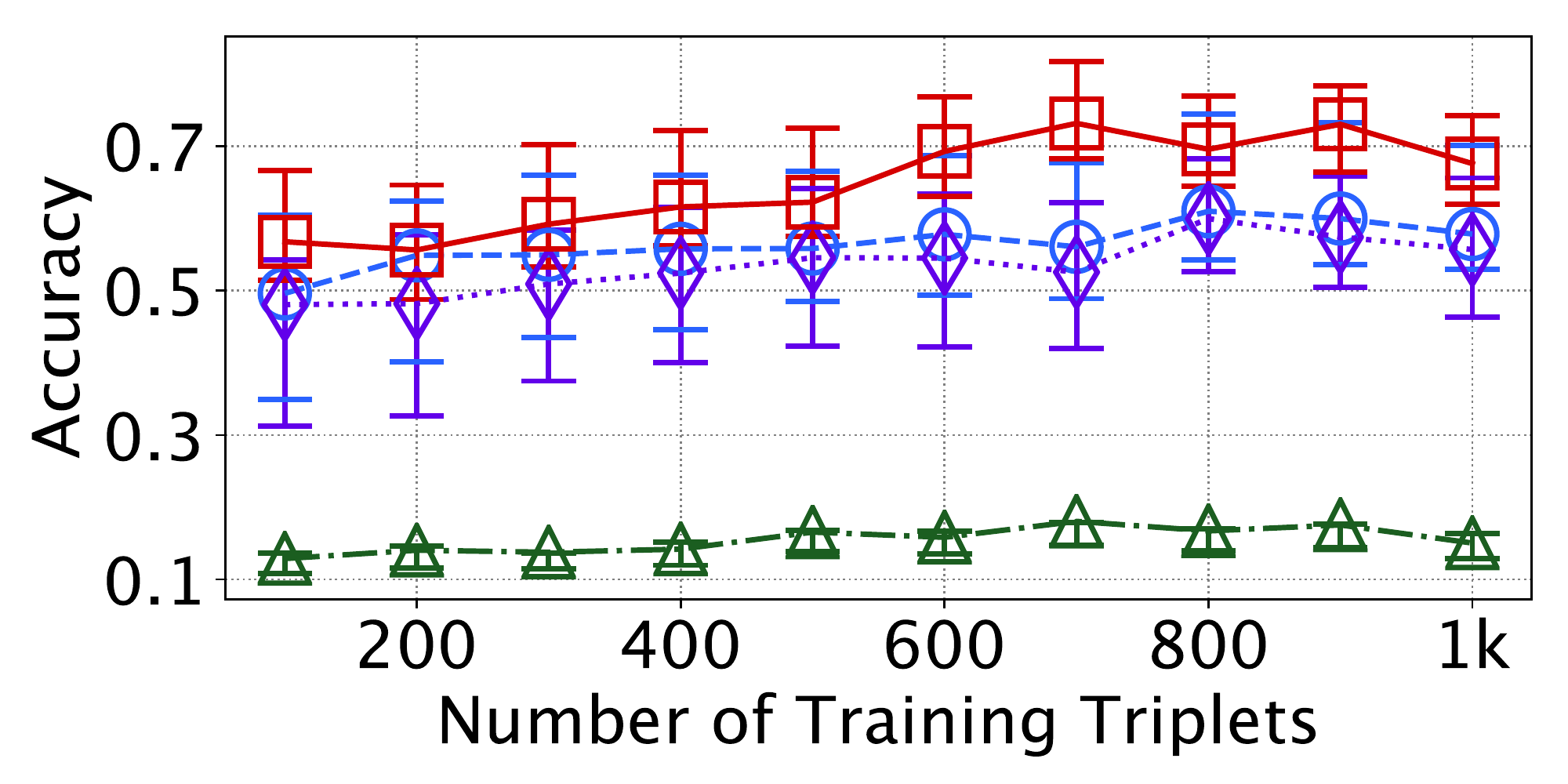}
        \vspace{-6mm}
        \caption{Dataset WS1}
        \label{fig-exp-mah-ws-1}
    \end{subfigure}
    \begin{subfigure}[t]{0.32\linewidth}
        \centering
        \includegraphics[width=\textwidth, page=2]{figures/experiments/0816/ws_manhattan_deviation}
        \vspace{-6mm}
        \caption{Dataset WS2}
        \label{fig-exp-mah-ws-2}
    \end{subfigure}
    \begin{subfigure}[t]{0.32\linewidth}
        \centering
        \includegraphics[width=\textwidth, page=3]{figures/experiments/0816/ws_manhattan_deviation}
        \vspace{-6mm}
        \caption{Dataset WS3}
        \label{fig-exp-mah-ws-3}
    \end{subfigure}
    
    \begin{subfigure}[t]{0.32\linewidth}
        \centering
        \includegraphics[width=\textwidth, page=4]{figures/experiments/0816/ws_manhattan_deviation}
        \vspace{-6mm}
        \caption{Dataset WS4}
        \label{fig-exp-mah-ws-4}
    \end{subfigure}
    \begin{subfigure}[t]{0.32\linewidth}
        \centering
        \includegraphics[width=\textwidth, page=5]{figures/experiments/0816/ws_manhattan_deviation}
        \vspace{-6mm}
        \caption{Dataset WS5}
        \label{fig-exp-mah-ws-5}
    \end{subfigure}
    \begin{subfigure}[t]{0.32\linewidth}
        \centering
        \includegraphics[width=\textwidth, page=6]{figures/experiments/0816/ws_manhattan_deviation}
        \vspace{-6mm}
        \caption{Dataset WS6}
        \label{fig-exp-mah-ws-6}
    \end{subfigure}
    \vspace{-2mm}
    \caption{Accuracy of the label prediction on Wikispeedia datasets using \textbf{Manhattan} distance function. }
    \vspace{-5mm}
    \label{fig-exp-mah-ws}
\end{figure*}

\subsubsection{Varying number of training triplets} 
Fig.~\ref{fig-exp-euc-ams} and~\ref{fig-exp-euc-ws} present the results where each setting has a fixed number of labels while the number of training triplets increases. Based on the experiment result figures, we have the following observations: 
\begin{itemize}
    \item As more triplets being used in the training process, the accuracy of one-shot learning keeps increasing with the trained \sysname network. Intuitively, with more examples demonstrated to the \sysnamens, it could better gain a better capability of generalization, even though the data instances used in one-shot learning are previously unseen. 
    \item Overfitting challenges the performance of all one-shot learning approaches. Although we use early stopping and $\ell_2$-regularization in all experiments, overfitting can still be challenging due to there is only one example per class in the one-shot learning. 
    \item \sysname can achieve better performance than other baseline methods when there are more possible classes. While \sysname maintains a stable performance outperforming state-of-the-art under various parameter settings, \sysname can achieve a better performance when the classification task become \textit{harder} with more possible class labels.  
\end{itemize}

\subsubsection{Observations using different datasets} Different from the synthetic datasets, the real-world applications often consist of diverse and noisy data instances. It is also interesting to examine the results using different real-world datasets. We find that the performance of \sysname remains superior when we use the twelve datasets sampled from two real-world applications.  

\subsubsection{Advantage of the \textit{end-to-end} model} Although it is possible to use attributed sequence embeddings~\cite{NAS-submission} with one-shot learning, the experiment results have proven that the performance of the end-to-end solution in this work is far superior to and more stable than \textsf{OLASEmb}. Specifically, the performance of our closest baseline method \textsf{OLASEmb} has varied more compared to all other methods.  Building an end-to-end model allows the back-propagation of gradient throughout all layers in the \sysname model. On the other hand, the two gradients in \textsf{OLASEmb}, \ie, the gradient in the model for generating attributed sequence embedding and the gradient in one-shot learning model, are independent and thus the parameters within this method cannot be better adjusted than our solution \sysnamens.

\subsubsection{Effect of different distance functions. } Recent work~\cite{bertinetto2016learning} has observed significant differences in performance when using different distance functions. Here, we substitute the Euclidean distance function with the Manhattan distance function to see the performance of all compared methods. We observe that the proposed \sysname model is capable of achieving the best results despite which one of the two distance functions is used. 

\section{Related Work}
\label{section-related}
\subsection{One-shot learning} 
One-shot learning, where the goal is to classify instances with only one example per class, has been the center of many applications~\cite{koch2015siamese,fei2006one,bertinetto2016learning,santoro2016meta,zheng2016towards,chao2016empirical,kopicki2016one,duan2017one,altae2017low,wu2018exploit}. It has been a useful approach to classification when the number of labeled instances is scarce. 
While conventional approaches to one-shot learning often involve Bayesian and shared probability densities~\cite{fei2003bayesian,miller2000learning}, 
recent works~\cite{koch2015siamese,bertinetto2016learning} take advantage of the feature learning capability of neural networks to further one-shot learning. 
The common objective of these tasks is to train a model so that the distance between instances from different classes is enlarged as much as possible~\cite{koch2015siamese,bertinetto2016learning}. Siamese network structure~\cite{koch2015siamese,bertinetto2016learning} is often used in such models, where two instances are taken as input, and the difference between them is learned. This example-based learning schema is flexible and has been applied to image classification tasks. 
However, these works focus on one-shot learning with only one type of data. In this paper, we further the state-of-the-art one-shot learning methods to learn from a more complicated data type (\ie, attributed sequence). 

\subsection{Distance Metric Learning}
Distance metric learning, where the goal is to learn a distance metric from pairs of similar and dissimilar examples, has been extensively studied~\cite{xing2003distance, yeung2007kernel, davis2007information, wang2011integrating, mignon2012pcca, koestinger2012large, cvpr-face-verify, mueller2016siamese, neculoiu2016learning,mlas-submission}. These tasks share a common objective of learning a distance metric, which could be used to reduce the distance between similar pairs of instances and increase the distance between dissimilar pairs of instances. Distance metric learning has shown its powerfulness in various tasks~\cite{xing2003distance, yeung2007kernel, davis2007information}. Many applications in various domains require 
 distance metric learning to achieve a good performance, such as identifying patient similarity in health informatics~\cite{wang2011integrating}, image recognition~\cite{koch2015siamese}, face verification~\cite{mignon2012pcca, koestinger2012large, cvpr-face-verify}, and sentence semantic similarity analysis~\cite{mueller2016siamese, neculoiu2016learning}. 
With the recent advancement in deep learning, distance metric learning has expanded to use various deep learning architectures to achieve its goal~\cite{koch2015siamese,mueller2016siamese,cvpr-face-verify}. 

\subsection{Deep learning} Deep learning has attracted a significant amount of research interest in recent years due to its capability of extracting features. Deep learning models, with a number of layers, are capable of learning features at various granularities. 
The capability of effective feature learning has advanced various research topics, including image recognition~\cite{karpathy2015deep,xu2015show} and sequence learning
~\cite{cho-EtAl:2014,sutskever2014sequence,xu2017decoupling,neculoiu2016learning,amas-submission}. 
It has also been applied in diverse problem domains, such as medical~\cite{sun2010localized} and traffic flow prediction~\cite{lv2015traffic}. Many of these applications involve only one type of data
~\cite{sutskever2014sequence,xu2017decoupling} while some applications make use of two types of data~\cite{karpathy2015deep,xu2015show}. However, none of these works has focused on this new data type of attributed sequence nor performing one-shot learning tasks on attributed sequences.

\section{Conclusion}
\label{section-conclusion}
In this paper, we study this new problem of one-shot learning for attributed sequences. We present the \sysname network design to tackle the challenges of utilizing this new data type in one-shot learning. \sysname incorporates two sub-networks, \cnet and \vnetns, that integrated into one structure together effectively learn the patterns hidden in this data type using only one example per class. \sysname uses this trained knowledge to generate labels for incoming unlabeled instances. Our experiments on real-world datasets demonstrate that \sysname on attributed sequences outperforms state-of-the-art one-shot learning methods.
\appendix[Gradients and Back-propagation in \sysname]
\label{app-a}
For the $m$-th layer in a fully connected neural network, we employ the following update functions:
\begin{equation}
    \begin{split}
        \frac{\partial \pmb{\upalpha}_m}{\partial \mathbf{W}_{m}} &= \pmb{\upalpha}_m\left(\mathds{1}-\pmb{\upalpha}_m\right)\pmb{\upalpha}_{m - 1} \\
        \frac{\partial \pmb{\upalpha}_m}{\partial \mathbf{b}_{m}} &= \pmb{\upalpha}_m\left(\mathds{1}-\pmb{\upalpha}_{m - 1}\right) 
    \end{split}
\end{equation}

Here we use three steps to explain how \sysname back-propagates the gradients. We use a $\delta_{\mu, \nu}$ function to simplify the equations with $\mu = \{\text{i, f, o}\}$ and $\nu = \{\text{i, f, o, c}\}$:
\begin{equation}
  \delta_{\mu, \nu}=\left\{
  \begin{array}{@{}ll@{}}
    1, & \text{if}\ \mu = \nu \\
    0, & \text{otherwise}
  \end{array}\right.
\end{equation} 
First, we have the following equations for $\mathbf{h}^{(t)}$ and $\mathbf{c}^{(t)}$: 
\begin{equation}
 \begin{split}
\frac{\partial   \mathbf{h}^{(t)}}{\partial \mathbf{W}_{\nu}} =   \frac{\partial \mathbf{o}^{(t)}}{\partial \mathbf{W}_{\nu}}\odot\tanh\big(&\mathbf{c}^{(t)}\big) +   \mathbf{o}^{(t)} \\ &\odot(1 - \tanh^2(\mathbf{c}^{(t)}))   \frac{\partial \mathbf{c}^{(t)}}{\partial \mathbf{W}_{\nu}}\\
\frac{\partial   \mathbf{h}^{(t)}}{\partial \mathbf{U}_{\nu}} =   \frac{\partial \mathbf{o}^{(t)}}{\partial \mathbf{U}_{\nu}}\odot\tanh(&\mathbf{c}^{(t)}) +   \mathbf{o}^{(t)} \\ &\odot(1 - \tanh^2(\mathbf{c}^{(t)}))   \frac{\partial \mathbf{c}^{(t)}}{\partial \mathbf{U}_{\nu}}\\
\frac{\partial   \mathbf{h}^{(t)}}{\partial \mathbf{b}_{\nu}} =   \frac{\partial \mathbf{o}^{(t)}}{\partial \mathbf{b}_{\nu}}\odot\tanh(&\mathbf{c}^{(t)}) +   \mathbf{o}^{(t)}\\ &\odot(1 - \tanh^2(\mathbf{c}^{(t)}))   \frac{\partial \mathbf{c}^{(t)}}{\partial \mathbf{b}_{\nu}}
\end{split}
\end{equation}
\vspace{-3mm}
\begin{equation}
  \begin{split}
\frac{\partial    \mathbf{c}^{(t)}}{\partial \mathbf{W}_{\nu}} =   \frac{\partial \mathbf{f}^{(t)}}{\partial \mathbf{W}_{\nu}}\odot\mathbf{c}^{(t-1)} &+   \mathbf{f}^{(t)}\odot  \frac{\partial \mathbf{c}^{(t-1)}}{\partial \mathbf{W}_{\nu}}   + \\  &\frac{\partial \mathbf{i}^{(t)}}{\partial \mathbf{W}_{\nu}}\odot\mathbf{g}^{(t)} +   \mathbf{i}^{(t)}\odot  \frac{\partial \mathbf{g}^{(t)}}{\partial \mathbf{W}_{\nu}} \\
\frac{\partial    \mathbf{c}^{(t)}}{\partial \mathbf{U}_{\nu}} =   \frac{\partial \mathbf{f}^{(t)}}{\partial \mathbf{U}_{\nu}}\odot\mathbf{c}^{(t-1)} &+   \mathbf{f}^{(t)}\odot  \frac{\partial \mathbf{c}^{(t-1)}}{\partial \mathbf{U}_{\nu}}   +   \\ &\frac{\partial \mathbf{i}^{(t)}}{\partial \mathbf{U}_{\nu}}\odot\mathbf{g}^{(t)} +   \mathbf{i}^{(t)}\odot  \frac{\partial \mathbf{g}^{(t)}}{\partial \mathbf{U}_{\nu}} \\
\frac{\partial    \mathbf{c}^{(t)}}{\partial \mathbf{b}_{\nu}} =   \frac{\partial \mathbf{f}^{(t)}}{\partial \mathbf{b}_{\nu}}\odot\mathbf{c}^{(t-1)} &+   \mathbf{f}^{(t)}\odot  \frac{\partial \mathbf{c}^{(t-1)}}{\partial \mathbf{b}_{\nu}}   +  \\  &\frac{\partial \mathbf{i}^{(t)}}{\partial \mathbf{b}_{\nu}}\odot\mathbf{g}^{(t)} +   \mathbf{i}^{(t)}\odot  \frac{\partial \mathbf{g}^{(t)}}{\partial \mathbf{b}_{\nu}}
\end{split}
\end{equation}

Then, we have the following equations for $\mathbf{i}^{(t)}, \mathbf{f}^{(t)}$ and $\mathbf{o}^{(t)}$:
\begin{equation}
\begin{split}
\frac{\partial  \Delta_\mu}{\partial \mathbf{W}_{\nu}} =  \Delta_\mu  (1 - \Delta_\mu) \vec{\alpha}^{(t)} \delta_{\mu, \nu} \\
\frac{\partial  \Delta_\mu}{\partial \mathbf{U}_{\nu}} =  \Delta_\mu  (1 -  \Delta_\mu) \mathbf{h}^{(t-1)} \delta_{\mu, \nu} \\
\frac{\partial  \Delta_\mu}{\partial \mathbf{b}_{\nu}} =  \Delta_\mu  (1 - \Delta_\mu) \delta_{\mu, \nu}
\end{split}
\end{equation}
where $ \Delta_{\text{i}} = \mathbf{i}^{(t)}$, $ \Delta_{\text{f}} = \mathbf{f}^{(t)}$ and $ \Delta_{\text{o}} = \mathbf{o}^{(t)}$. 

Finally, we have the gradients for $\mathbf{g}^{(t)}$ as:
\begin{equation}
\begin{split}
\frac{\partial \mathbf{g}^{(t)}}{\partial \mathbf{W}_{\nu}} =   (1 -  (\mathbf{g}^{(t)}) ^2)\vec{\alpha}^{(t)} \delta_{c, \nu} \\[1pt]
\frac{\partial \mathbf{g}^{(t)}}{\partial \mathbf{U}_{\nu}} = (1 - (\mathbf{g}^{(t)}) ^2)\mathbf{h}^{(t-1)} \delta_{c, \nu} \\[1pt]
\frac{\partial \mathbf{g}^{(t)}}{\partial \mathbf{b}_{\nu}} =  (1 -  (\mathbf{g}^{(t)}) ^2) \delta_{c, \nu}
\end{split}
\end{equation}

\bibliographystyle{IEEEtran}
\bibliography{sections/references_full}

\end{document}